\documentclass[letterpaper, 10 pt, journal, twoside]{IEEEtran}

\usepackage{cuted}

\usepackage{graphics} 
\usepackage{epsfig} 
\usepackage{leftindex}
\usepackage{amsmath} 
\usepackage{bm}
\usepackage{amssymb}  
\usepackage{algorithm}
\usepackage[noend]{algorithmic}
\usepackage{xcolor}
\usepackage{graphicx,booktabs,array}
\usepackage[hidelinks]{hyperref}
\usepackage{caption}
\usepackage{subcaption}

\usepackage{physics}
\usepackage{soul} 
\usepackage{color}

\newcommand{\mc}[1]{\mathcal{#1}}

\title{\textbf{DANCeRS}: A \textbf{D}istributed \textbf{A}lgorithm for \textbf{N}egotiating \textbf{C}ons\textbf{e}nsus in \textbf{R}obot \textbf{S}warms with Gaussian Belief Propagation}

\begin{document}

\author{Aalok Patwardhan$^{1}$ and Andrew J. Davison$^{1}$
\thanks{$^{1}$Aalok Patwardhan and Andrew J. Davison are with the Dyson Robotics Lab and the Department of Computing, Imperial College London 
{\tt\footnotesize [a.patwardhan21,a.davison]@imperial.ac.uk}}%
        }
\maketitle

\begin{abstract}
Robot swarms require cohesive collective behaviour to address diverse challenges, including shape formation and decision-making. Existing approaches often treat consensus in discrete and continuous decision spaces as distinct problems. We present DANCeRS, a unified, distributed algorithm leveraging Gaussian Belief Propagation (GBP) to achieve consensus in both domains. By representing a swarm as a factor graph our method ensures scalability and robustness in dynamic environments, relying on purely peer-to-peer message passing.
We demonstrate the effectiveness of our general framework through two applications where agents in a swarm must achieve consensus on global behaviour whilst relying on local communication. In the first, robots must perform path planning and collision avoidance to create shape formations. In the second, we show how the same framework can be used by a group of robots to form a consensus over a set of discrete decisions. Experimental results highlight our method's scalability and efficiency compared to recent approaches to these problems making it a promising solution for multi-robot systems requiring distributed consensus. We encourage the reader to see the supplementary video demo.

\end{abstract}

\section{INTRODUCTION}

In nature swarms such as bird murmurations or schools of fish consist of simple agents that communicate locally and exhibit collective behaviour, often requiring agreement on shared global parameters like velocity or orientation. Similarly, robot swarms must act cohesively, whether for task allocation, arrangement, or coordinated behaviours like exploration or aggregation \cite{brambilla_review_swarm}. This requires each robot to make individual decisions while aligning with the swarm's global objectives.

Centralised systems offer precise coordination by delegating all computations to a single external entity. However, such approaches lack scalability and are prone to single points of failure. In contrast, decentralised systems rely on distributed computation and local communication, enabling agents to share information and iteratively refine their decisions based on their neighbours' states. For discrete decision spaces, agents often share probability distributions over the available choices and update their beliefs until consensus is reached \cite{consensus_ECA}. These methods are effective for a swarm deciding between distinct behaviours like exploration \cite{Patwardhan:GBPStack}, aggregation, or task allocation, commonly framed as best-of-N problems \cite{valentini_bestofn}.
\begin{figure}[t]
    \centering
    \includegraphics[width=0.9\linewidth]{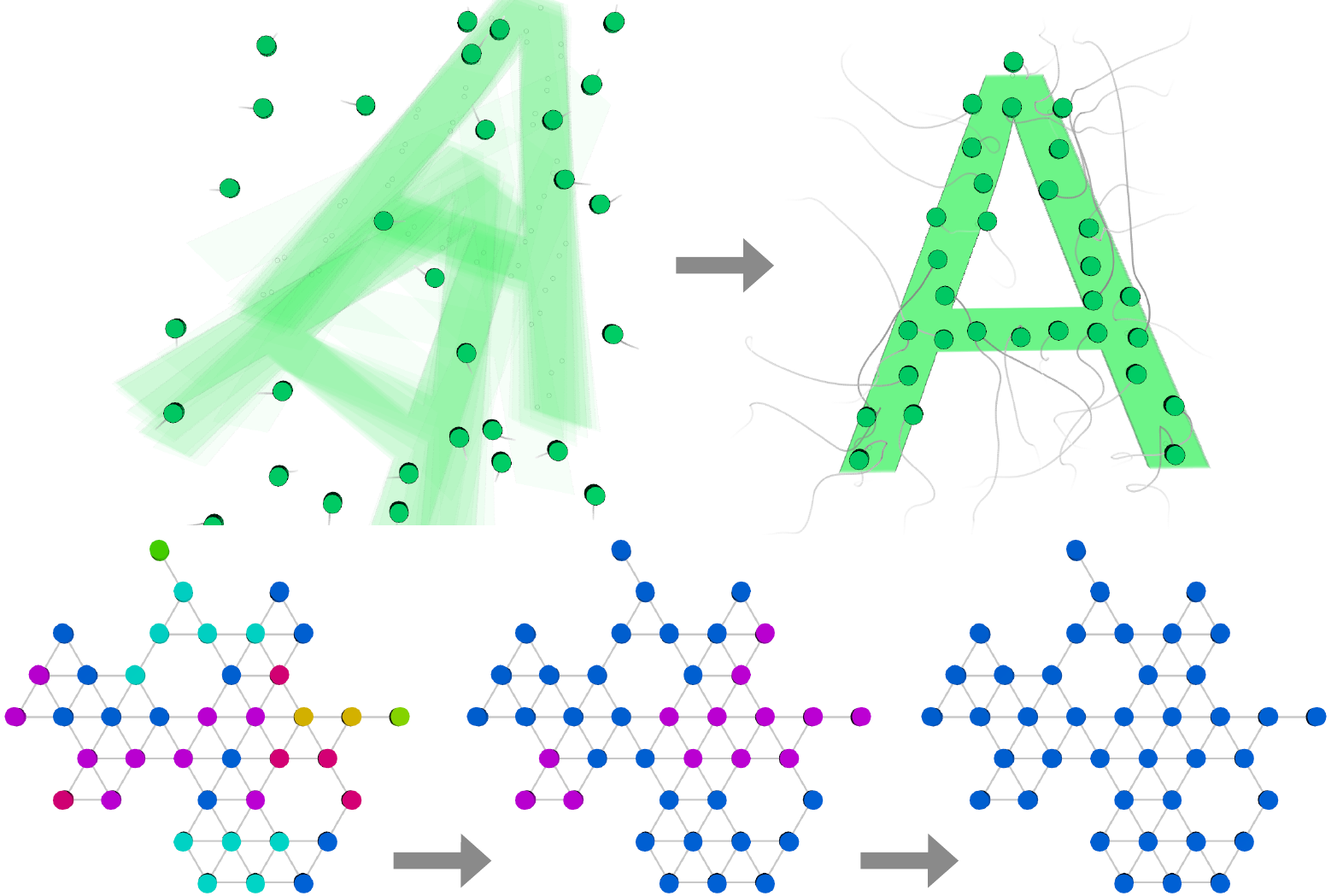}
    \caption{(Top) Robots jointly plan paths to arrange into the `A' shape, achieving consensus over its pose. (Bottom) Robots use the \emph{same} algorithm to achieve consensus over a discrete set of decisions (colours) through local communication.}
    \label{fig:teaser}
\end{figure}

For continuous spaces, consensus is typically achieved by iteratively updating shared variables, such as positions or orientations, based on differences between neighbouring agents' states. This allows swarms to converge toward global parameters collaboratively, even with limited local communication. Such approaches have been used for shape formation \cite{meanshift}, where robots must agree on the position and orientation of a shape for coordinated motion. These distributed methods provide scalability and robustness, enabling swarms to function in dynamic environments without centralised control.

Despite significant progress, the literature typically treats consensus over discrete decision spaces and continuous parameter spaces as separate problems. In this work, we propose a unified framework for achieving consensus in both domains. By modelling the swarm as a sparsely connected factor graph and using Gaussian Belief Propagation (GBP) for distributed inference, our method enables robots to reach consensus efficiently and scalably. This fully distributed approach ensures robustness to increasing swarm size and dynamic environments, addressing both discrete and continuous decision-making in a single cohesive framework.

We demonstrate our general framework in two example applications of joint path planning for shape formation and collective decision-making over a set of discrete options. Our contributions are:
\begin{itemize} 
\item A unified, distributed framework for consensus in both discrete and continuous decision spaces. 
\item The extension of GBP path planning for non-holonomic unicycle dynamics in shape formation. 
\item A novel distributed shape formation method for discrete target assignment, demonstrated with scalability and efficiency in experiments.
\item The first application of GBP for discrete consensus in robot swarms.
\end{itemize}
\section{RELATED WORK}
Achieving global behaviour in robot swarms relies on decision-making frameworks that balance scalability and robustness. While centralised methods are limited by scalability and single points of failure, decentralised approaches enable collective behaviour through local interactions, whilst addressing challenges such as conflict resolution. The literature focuses on two key problems: consensus over a set of discrete options and over continuous parameter spaces.

In discrete consensus problems, each agent in the swarm must select one option from a finite set based on local communication with its neighbours. This problem, often referred to as the ``best-of-$N$'' problem has been widely studied particularly for $N\mathord{=}2$ options where a swarm estimates the majority value in the environment \cite{Shan2021:Discrete, siemensma:bayesian}. For larger $N$, probabilistic approaches involve agents maintaining and sharing a probability distribution over the set of options. In \cite{consensus_PCA} agents update their internal beliefs based on local communication and knowledge of the local consensus group (agents sharing the same decision) whereas in \cite{consensus_ECA} agents share decision IDs and levels of certainty of their beliefs based on probabilistic entropy. Other popular distributed approaches include voter methods where robots use information from their neighbours and choose a random decision \cite{gross_consensus}, the majority decision \cite{majority} or the most frequent recently observed decision \cite{k-unamity}. 
Another approach to distributed decision-making is opinion dynamics, which models how agents adjust their opinions through local interactions. Unlike traditional consensus models, it allows for complex social dynamics, including agreement, disagreement, and opinion clustering. The nonlinear model in \cite{Bizyaeva:OpinionDynamics} introduces tunable sensitivity, enabling flexible opinion transitions. While this provides insight into how opinions evolve in networks, such models do not always ensure that all agents reach the same decision, which is often desirable in multi-agent coordination tasks.

Continuous consensus problems involve the swarm negotiating their beliefs about continuous global parameters, such as target locations in search-and-rescue or source-seeking tasks, or the position and orientation of a reference frame for shape formation. Distributed consensus algorithms typically rely on robots effectively averaging their neighbours' beliefs for formation control \cite{savkin}, obstacle avoidance \cite{afdila} and saliency detection \cite{alhafnawi}. Recent work applied the mean-shift algorithm to shape formation tasks \cite{meanshift}, where robots negotiated on the pose of the shape, and were able to form and explore complex patterns defined as an artificial potential field. This work was extended in \cite{meanshift_targetfree} to consider formations consisting of distinct points. These mean-shift approaches iteratively adjust robots' velocities to align with local density maxima, enabling collision avoidance and dynamic role assignment, although they are designed for shapes with one continuous connected component.
Other approaches, such as local task swapping, have been proposed for distributed shape formation \cite{wang2020localtaskswap}. In this method, robots iteratively negotiate position assignments by swapping roles, deciding whether to act as anchors or clients based on local priorities. A hop-count strategy propagates goal selection, ensuring collision-free movement in a discretised grid-world. However, this approach assumes robots have already reached consensus on a shared global reference frame. By contrast, we seek to addresses both shape formation and reference frame alignment simultaneously, removing the need for task negotiation based on roles or priority.

Despite being treated as two distinct problems in the literature, real-world tasks often require swarms to achieve consensus in both discrete and continuous decision spaces simultaneously. In our work, we propose a unified framework capable of addressing both types of consensus problems. Our approach builds on Gaussian Belief Propagation (GBP), a distributed algorithm for factor graph inference, which has demonstrated strong performance in dynamically changing graph topologies. GBP is asynchronous, fully distributed, and relies on peer-to-peer message passing between nodes in the graph.
Our previous works \cite{Patwardhan:GBPStack}, \cite{Patwardhan:GBPPlanner} applied GBP to distributed multi-robot path planning and information aggregation but were limited to Euclidean spaces and holonomic motion models. More recently, GBP has also been used for consensus in 2D spaces \cite{GBP_simon}, but without extending to Lie groups, restricting its ability to handle more complex transformations.

In this work we introduce a general GBP-based framework for consensus over Lie groups, enabling robots to negotiate over more complex spaces beyond simple Euclidean variables. Unlike \cite{Patwardhan:GBPStack}, where consensus was achieved through implicit measurement fusion, we formulate it as an explicit negotiation process. Additionally, we extend GBP to discrete decision spaces, broadening its applicability beyond continuous estimation. We also propose a novel shape formation algorithm, leveraging GBP for distributed coordination, and incorporate a non-holonomic motion model, making our approach more realistic than prior work for practical robotic systems.
\section{BACKGROUND}

\subsection{Gaussian Belief Propagation (GBP) for Factor Graphs}
Probabilistic Graphical Models (PGMs) are a powerful framework for representing sparse optimisation problems, where a set of variables $\bm{\mathrm{X}}$ are connected by constraints. The joint function $p(\bm{\mathrm{X}})$ can be factorised into factors $f_s$ that encapsulate constraints or cost functions:
\begin{equation}
\label{eqn:factorprod}
p(\bm{\mathrm{X}}) = \prod_s f_s(\bm{\mathrm{X}}_s)
~.
\end{equation}
When the factors and variables take on Gaussian distributions we can use Gaussian Belief Propagation (GBP) to perform marginal inference on the resulting factor graph. In GBP variables and factors are nodes which perform purely local node-to-node message passing and update their beliefs. At any time, the current marginal beliefs for the variables can be obtained.
We represent a Gaussian distribution in information form as:
\begin{equation}
    \mc{N}(\bm{\mathrm{X}}; \bm{\mu}, \bm{\Sigma}) = \mc{N}^{-1}(\bm{\mathrm{X}}; \bm{\eta}, \bm{\Lambda})~,
\end{equation}
where  $\bm{\Lambda} = \bm{\Sigma}^{-1}$ and $\bm{\eta} = \bm{\Lambda} \bm{\mu}$ are the precision matrix and information vector respectively.
In GBP, variables in $\bm{\mathrm{X}}_s$ are assumed to be Gaussian: each variable $\bm{\mathrm{x}}_k$ has a belief  $b(\bm{\mathrm{x}}_k) = \mc{N}^{-1}(\bm{\mathrm{x}}_k; \bm{\eta}_k, \bm{\Lambda}_k)$.
A Gaussian factor $f_s(\bm{\mathrm{X}}_s)$ can be any (non-linear) function that connects variables $\bm{\mathrm{X}}_s$ of the form:
\begin{equation}
\label{equ:generalfactor}
f_s(\bm{\mathrm{X}}_s) \propto e^{-\frac{1}{2} \bm{r}^\top \bm{\Lambda}_s \bm{r}  }
~,
\end{equation}
where the residual $\bm{r} = \bm{z}_s - \bm{h}_s(\bm{\mathrm{X}}_s)$, $\bm{h}_s(\bm{\mathrm{X}}_s)$ represents the linearised measurement function or constraint that the factor represents, $\bm{z}_s$ is its observed value and $\bm{\Lambda}_s$ is the precision of the constraint.

One iteration of GBP consists of three steps involving node-to-node message passing which we outline briefly here; see \cite{davison2019futuremapping} for a more in-depth analysis and derivation.

\subsubsection{Factor to Variable Message}  
A message from a factor $f_j$ to one of its connected variables $\bm{\mathrm{x}}_k$ is computed by marginalizing the product of the factor's potential (the exponent in Equation \ref{equ:generalfactor}) and all incoming variable to factor messages, excluding the message from $\bm{\mathrm{x}}_k$.

\subsubsection{Variable Belief Update}
\label{sec:variable_belief_update}
A variable $\bm{\mathrm{x}}_k$ updates its belief by taking the product of all incoming messages from its connected factors.

\subsubsection{Variable to Factor Message}
A message from a variable $\bm{\mathrm{x}}_k$ to a connected factor $f_j$ is the product of all incoming factor to variable messages except the message from $f_j$.

\subsection{GBP using Lie Theory}
Gaussian Belief Propagation (GBP) is commonly used in Euclidean spaces, where variables are represented as vectors. However, many robotics problems require optimisation over Lie groups, which provide a natural way to represent transformations, orientations, and motions. Lie groups are continuous mathematical structures that generalize Euclidean spaces while preserving properties like smoothness and group operations. They are widely used in robotics to represent states such as positions, rotations, and velocities in a way that respects the underlying geometry.

In consensus problems, robots must agree on shared parameters such as position, orientation, or categorical decisions. While averaging works well for simple Euclidean quantities, it fails for rotations and transformations. For example, in the $SO(2)$ Lie group, averaging two angles near 0° and 360° incorrectly suggests 180°, rather than recognising their proximity. Lie theory allows optimisation to be performed on the correct mathematical space, ensuring accurate consensus.

GBP operates by passing Gaussian-distributed messages between variables in a factor graph. In Euclidean settings, these messages contain a mean vector and precision matrix, but for Lie groups, they must be transformed into the tangent space of the current belief, updated, and mapped back using exponential and logarithmic maps. This ensures that information is propagated correctly without distorting the underlying geometry. For a detailed derivation and further insights, see \cite{murai2022robot}, \cite{microlie}. Our method provides a general framework for optimisation on Lie groups, enabling applications in trajectory planning, state estimation, and control. In this work, we demonstrate its use for multi-robot consensus over poses and discrete decisions.
\section{METHOD}
We consider a swarm of $N$ robots moving in $\mathbb{R}^2$ each having a communications radius $r_C$. At any time $t>0$ they form a sparsely connected undirected graph $\mc{G}=(\mc{V},\mc{E})$ where $\mc{V}=\{\mc{V}_i : i \in [0, ..., N-1]\}$ is the set of robots, and $\mc{E}\subseteq\mc{V}\times\mc{V}$ is the set of edges denoting inter-robot connections. An edge $\mc{E}_{ij}$ between robot $i$ and $j$ exists if at time $t$ if $\norm{\boldsymbol{x}_i - \boldsymbol{x}_j} < r_C$. 

\begin{figure}[h]
    \centering
    \includegraphics[width=0.8\linewidth]{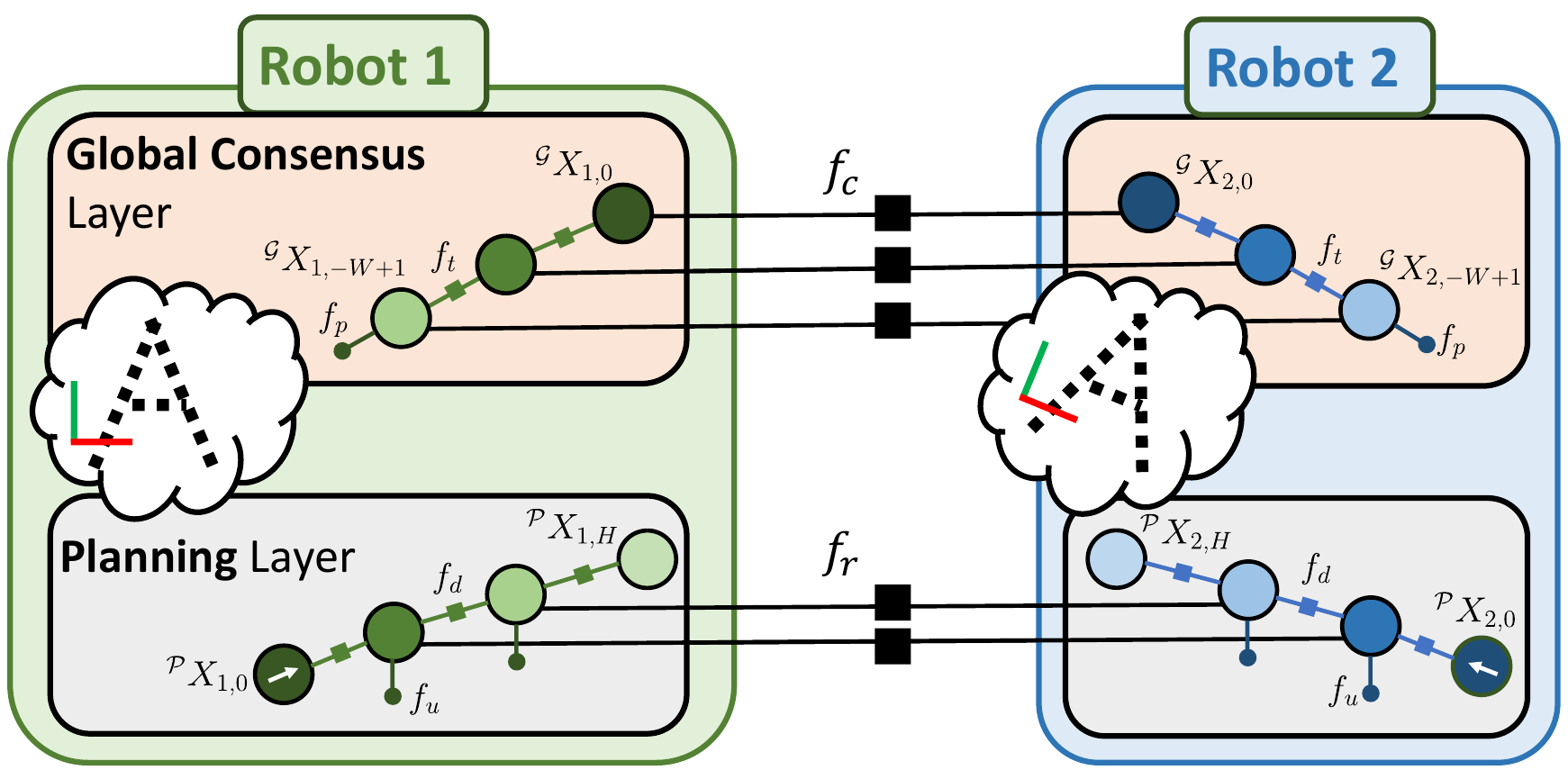}
    \caption{Factor graph stack of each robot. Local GBP iterations enable global consensus between the robots, visualised here as the formation parameters for the letter A.}
    \label{fig:gbpstack}
\end{figure}

The robots must solve multi-faceted problems; they must perform optimisation in path planning and collision avoidance, but also form consensus over some shared global information. Similar to the approach from \cite{Patwardhan:GBPStack}, each robot holds a two-layered factor graph stack as shown in Figure \ref{fig:gbpstack} where each layer of the stack represents a particular aspect of the optimisation problem -- the Planning and Global Consensus layers.

As ours is a purely distributed algorithm we take inspiration from \cite{murai2022robot}; each robot maintains a ``webpage'' of information containing the outgoing messages to the factors and variables in the GBP stacks of each of its connected neighbours. Each robot follows Algorithm \ref{algo:general}. 

The parametrisation of variables in the Global Consensus layer depends on the specific application to be considered. To demonstrate our general framework which works for many of the common Lie Groups, we perform two types of experiments where the robots must reach consensus over some shared global information.

In the first experiment we choose shape formation consensus as a proxy problem where the $SE(2)$ group can be used to represent the position and orientation of a target formation shape.
In the second we examine how the same framework can be used for performing negotiation over the continuous $\mathbb{R}^M$ group, and can be used to perform decision-making and negotiation over a set of discrete options.

We now describe the general form of the two factor graph layers as depicted in Figure \ref{fig:gbpstack}. In general we describe a factor $f_s$ with its measurement function $h_s$ and its strength $\boldsymbol{\sigma}_s$ such that $\boldsymbol{\Lambda}_s=diag(\boldsymbol{\sigma}_s^{-2})$.

\subsection{Global Consensus Layer}
In this layer robots may share information about their interpretation of the global parameters and negotiate to collectively form a consensus.

Let $\chi$ represent the global parameter that all robots must agree upon. For example, this may represent a shared behaviour to be followed, or alternatively a common reference frame to be agreed upon by the robots.
Each robot holds variables $\leftindex^{\mc{G}}X \in \mc{M}$ representing the robot's interpretation of the global parameters. In our generalised framework $\mc{M}$ may be one of the common Lie groups such as $\mathbb{R}^M, SO(2), SO(3), SE(2), SE(3)$.

\subsubsection{Consensus Domain}
The GBP variable $\leftindex^{\mc{G}}{X}_{i}$ in this layer represents robot $i$'s belief about the global parameter $\chi$, and consensus is achieved over the continuous domain $\mc{M}$.

When applying our method to problems involving consensus over a discrete set of options, we choose $\mc{M}=\mathbb{R}^1$ to represent the continuous form of the discrete decision space.
We define the quantisation function $\gamma: [0, 1] \to \{0, 1, \ldots, N_D-1\}$ and its inverse $\gamma^{-1}$ as
\begin{eqnarray}
\gamma(x) =& \left\lfloor N_D \cdot x \right\rfloor \\
\gamma^{-1}(k) =& \frac{k}{N_D} \quad \text{for} \quad k = 0, \ldots, N_D-1
\end{eqnarray}
respectively where \( N_D \in \mathbb{Z}_{>0} \) is the total number of discrete intervals.
Transforming the discrete decision space into a continuous space allows the use of GBP for iterative negotiation, enabling the swarm to converge onto a shared common decision. It should be noted that the quantisation function is used simply to extract the robot's discrete decision, and is not part of the consensus algorithm.

\subsubsection{Prior Factor}
Robot $i$ has a prior belief about the global parameter represented by a factor of strength $\boldsymbol{\sigma}_p$ and form
\begin{equation}
f_p: h_p(\leftindex^{\mc{G}}X_i) = \leftindex^{\mc{G}}X_i~\ominus~\leftindex^{\mc{G}}x_i^0
\end{equation}
where $\leftindex^{\mc{G}}x^0$ is the initial belief of the formation parameters, and $\ominus$ is the right--minus action on the manifold.

\subsubsection{Inter-robot Consensus Factor}
When two robots are within communication range of each other, inter-robot factors are created with strength $\boldsymbol{\sigma}_c$ and take the form
\begin{equation}
\label{eqn:interrobot_factor}
f_c: h_c(\leftindex^{\mc{G}}X_i, \leftindex^{\mc{G}}X_j) = \leftindex^{\mc{G}}X_i \ominus \leftindex^{\mc{G}}X_j = Log(\leftindex^{\mc{G}}X_j^{-1} \cdot \leftindex^{\mc{G}}X_i)~.
\end{equation}

\subsubsection{Temporal Factors and Variables}
The resulting factor graph is dynamic as robots may leave and re-join a local group at any time. When robots leave a group, they delete their existing inter-robot factors -- as GBP variables are memory-less, robots would forget the locally converged belief of the global parameter. Hence, robots maintain a sliding window of $W$ variables $\leftindex^{\mc{G}}X_{i,-k} : k \in [0,W-1]$ in this layer, which are connected with temporal factors $f_t(\leftindex^{\mc{G}}X_{i,-k-1},\leftindex^{\mc{G}}X_{i,-k})$ having the same form as the inter-robot factor in Equation \ref{eqn:interrobot_factor} with a strength $\boldsymbol{\sigma_t}$.
The sliding window is updated every $T_S$ timesteps. The oldest variable $\leftindex^{\mc{G}}X_{i,-W+1}$ along with its connected factors is deleted. A new prior factor $f_{p}$ is generated at the now oldest variable $\leftindex^{\mc{G}}X_{i,-W+2}$ and initialised with the message from factor $f_t(\leftindex^{\mc{G}}X_{i,-W+1}, \leftindex^{\mc{G}}X_{i,-W+2})$. This prior factor represents the marginalised information from the deleted portion of the factor graph.

The resulting effect is that a robot disconnecting from a local group maintains its previous belief mean, although its covariance weakens over the sliding window.

\subsection{Path Planning Layer}
Optimisation in this layer involves robots planning an efficient path in a short forward time window of $H+1$ timesteps, and creating inter-robot factors between themselves and their neighbours for collision avoidance.

A noise-on-acceleration model is used for the robot dynamics such that the robot state are parametrised as 
\begin{equation}
    \leftindex^{\mc{P}}X=[\boldsymbol{p}^\top, \boldsymbol{\dot{p}}^\top]^\top = [x, y, \theta, \dot{x}, \dot{y}, \dot{\theta}]^\top~.
\end{equation}

We build upon the work presented in \cite{Patwardhan:GBPPlanner} where robots with holonomic dynamics models planned paths towards fixed goals. These paths consisted of the robots' planned states $^\mc{P}\mathbf{X}_k$ for $k \in \left[0, H\right]$ which are the optimisation variables, connected to each other through dynamics factors $f_d$ representing a constant velocity model constraint.
In this work however we propose a new unicycle model factor $f_u$, enabling modelling of non-holonomic motion, and improvements over the previous work, which we detail here.
\subsubsection{Unicycle Model Factor}
This factor with strength $\sigma_u$ is added to each variable in the layer and takes the form:
\begin{equation}
f_u : h_u(\leftindex^{\mc{P}}X_k) = \dot{x}\cos(\theta) - \dot{y}\sin(\theta)
\end{equation}
As factors represent cost functions, optimisation drives the value of the measurement function of this factor $h_u(\leftindex^{\mc{P}}X_k)$ to $0$. This encourages the robot velocity at any time $k$ to be in the direction of the robot's heading.

\subsubsection{Collision Avoidance Factor}
We also introduce a new inter-robot collision avoidance factor between the variables of robot $i$ and $j$ which has strength $\sigma_r$ and takes the form:
\begin{equation}
f_r : h_r(\leftindex^{\mc{P}}X_{k, i}, \leftindex^{\mc{P}}X_{k, j}) = \exp{-\frac{\left| \boldsymbol{x}_{k, i} - \boldsymbol{x}_{k, j} \right|}{d_{min}}}
\end{equation}
where $d$ is the distance between the two robot planned paths at timestep $k$, and $d_{min}$ is a minimal separation distance between the robots.
This factor has a smoother measurement function compared to the equivalent factor in \cite{Patwardhan:GBPPlanner} and so a more well-defined Jacobian.

\subsubsection{Horizon Update}
As per Algorithm \ref{algo:general}, at every time step after $N_I$ iterations of GBP the current and horizon variables of the planned path are propagated forward in time. Unlike previous work, we move the horizon state $\boldsymbol{p}_{H}$ towards an application-specific goal location $\boldsymbol{g}$ which can be time-varying:
\begin{eqnarray}
    &\boldsymbol{p}_{H} \leftarrow \boldsymbol{p}_{H} + \boldsymbol{\dot{p}}_{H} \Delta t \\
    &\boldsymbol{\dot{p}}_{H} = \begin{bmatrix}
    \min\left(v_{\text{max}}, \frac{\|\boldsymbol{d}\|}{\Delta t} \right) \cdot \frac{\boldsymbol{d}}{\|\boldsymbol{d}\|} \\ \min\left(\omega_{\text{max}}, \frac{\abs{\angle\boldsymbol{d} - \boldsymbol{p}_{H, \theta}}}{\Delta t} \right) \cdot sign(\angle\boldsymbol{d} - \boldsymbol{p}_{H, \theta})
    \end{bmatrix} \\
    &\boldsymbol{d} = \begin{bmatrix} \boldsymbol{g}_x - \boldsymbol{p}_{H,x} & \boldsymbol{g}_y - \boldsymbol{p}_{H,y} \end{bmatrix}^\top
\end{eqnarray}

If the specific application requires the robots to randomly explore the environment, a new $\boldsymbol{g}$ is selected when the robot's horizon state is sufficiently near it. Alternatively if the swarm is to create a shape formation, $\boldsymbol{g}$ can be selected using a nearest-neighbour search over potential locations using a novel distance-occupancy based heuristic we propose in section \ref{sec:shape_formation}.

\begin{algorithm}[t]
\footnotesize
\caption{For each robot $i$}\label{algo:general}
\begin{algorithmic}[1]
 \STATE Initialise GBP stack layers $\mathcal{P,G}$
 \WHILE{$t<T_{max}$}
 \STATE \textit{Let $\mc{N}(i) = \{j\ |\ ||{\boldsymbol{p}_{i,0}-\boldsymbol{p}_{j,0}}|| < r_C  \}$ be the set of robots within the communication radius of $i$.}\\
 \STATE \textit{Let $\mc{C}(i)$ be the set of robots connected to $i$.}\\
 \FOR {Newly observed robot $j \in \mc{N}(i) \backslash \mc{C}(i)$}
 \STATE {Create inter-robot factors $f_c, f_r$.}
 \ENDFOR
 \FOR {Out-of-range robot $j \in \mc{C}(i) \backslash \mc{N}(i)$}
 \STATE {Delete inter-robot factors $f_c, f_r$.}
 \ENDFOR
 \FOR {layer $\mathcal{L}$ in GBP Stack}
 \STATE Broadcast inter-robot messages.
 \STATE Perform $N_I$ iterations of GBP.
 \ENDFOR
 \STATE Update $^\mc{P}\boldsymbol{X}_0$ and $^\mc{P}\boldsymbol{X}_{H}$ by $\Delta t$.
 \STATE Create new information layer variables $^\mc{G}\boldsymbol{X}_{k}$ with temporal factors $f_t$.
 \ENDWHILE
\end{algorithmic} 
\end{algorithm}
\section{APPLICATIONS AND SIMULATIONS}
\subsection{Path Planning and Consensus for Shape Formation}
\label{sec:shape_formation}
In this problem a swarm of robots with a limited communication radius $r_C$ must perform path planning to efficiently arrange themselves into a shape formation.
Each robot has knowledge of the shape of the formation but must negotiate on the formation parameters (position and orientation) with the other robots.
The shape of the formation is stored as a set of points in $\mathbb{R}^2$ with given minimum spacing $r_S$, and robots select the nearest point as the path planning goal $\boldsymbol{g}$ using a heuristic depending on distance and occupancy based on local information.

The shape of the formation is defined as a set of $N_F$ points $q = \begin{bmatrix}q_x & q_y\end{bmatrix}^\top \in \mathbb{R}^2$ known by all robots in a canonical coordinate frame of reference. We consider $N_F=N_R$, and at any time robots may use their current belief about the formation parameters $\leftindex^{\mc{G}}{X}_{i,0} = Exp \left(\begin{bmatrix}x_{i,0} & y_{i,0} & \theta_{i,0}\end{bmatrix}\right)$ to transform a point $\boldsymbol{p}$ in the canonical frame to the global frame through the group action
$\boldsymbol{p}' = \leftindex^{\mc{G}}{X}_{i,0} \circ \boldsymbol{p}$.
The transformation from global to canonical frame can be done with the group inverse $\leftindex^{\mc{G}}{X}_{i,0}^{-1}$.
In order to demonstrate our generalised framework, we opt to use the $SE(2)$ group to represent the formation parameters, although alternative representations of position and orientation such as the decoupled bundle of $\left<\mathbb{R}^2, SO(2) \right>$ could also be used; see \cite{microlie} for further details.

\subsubsection{Occupancy Weighting}
Robots must select points that are unoccupied by other robots, however this is hindered by the limited communications range. In this scenario robots must be able to explore within the shape. The set of 2D formation points (FPs) $Q_i$ is augmented with an extra value $\tau$ for each point which we refer to as the `occupancy weighting' of the point.
We denote this augmented set of points as 
\begin{equation}
    \Tilde{Q}_i = \left\{[q_{i,m}^\top~ \tau_{i,m}]^\top~ \forall~ q_{i,m} \in Q_i \right\}~.
\end{equation}

Each robot stores $\Tilde{Q}_i$ as a KD-Tree data structure which enables efficient and fast nearest-neighbour searching with Euclidean distance metrics.
The occupancy weighting (OW) dimension of an FP $\Tilde{q}_{i,m}(\tau)$ reflects how recently the point was seen as being occupied by a neighbouring robot.
Robots follow Algorithm \ref{algo:formation} to select an optimal FP for the path planning goal $\boldsymbol{g}$, and can make use of the most recent beliefs about the positions of neighbouring robots from the path planning layer. For a robot $i$, if any neighbouring robots are near (within a range $r_N$) to an FP $q_{i,m}$, the OW value of the point $\Tilde{q}_{i,m}(\tau)$ is set to an arbitrarily high value of $\tau_0$. For the remaining FPs within a distance $r_C$ of robot $i$ are set with $\tau_{i,m}=0$ in the OW dimension as there are no neighbours occupying them.
For all other FPs, as further information about the occupancy is unavailable their OW values are decremented by $1$. This OW `decay' indicates that they may still be occupied.

\subsubsection{Goal selection}
\begin{figure}[h]
    \centering
    \includegraphics[width=0.7\linewidth]{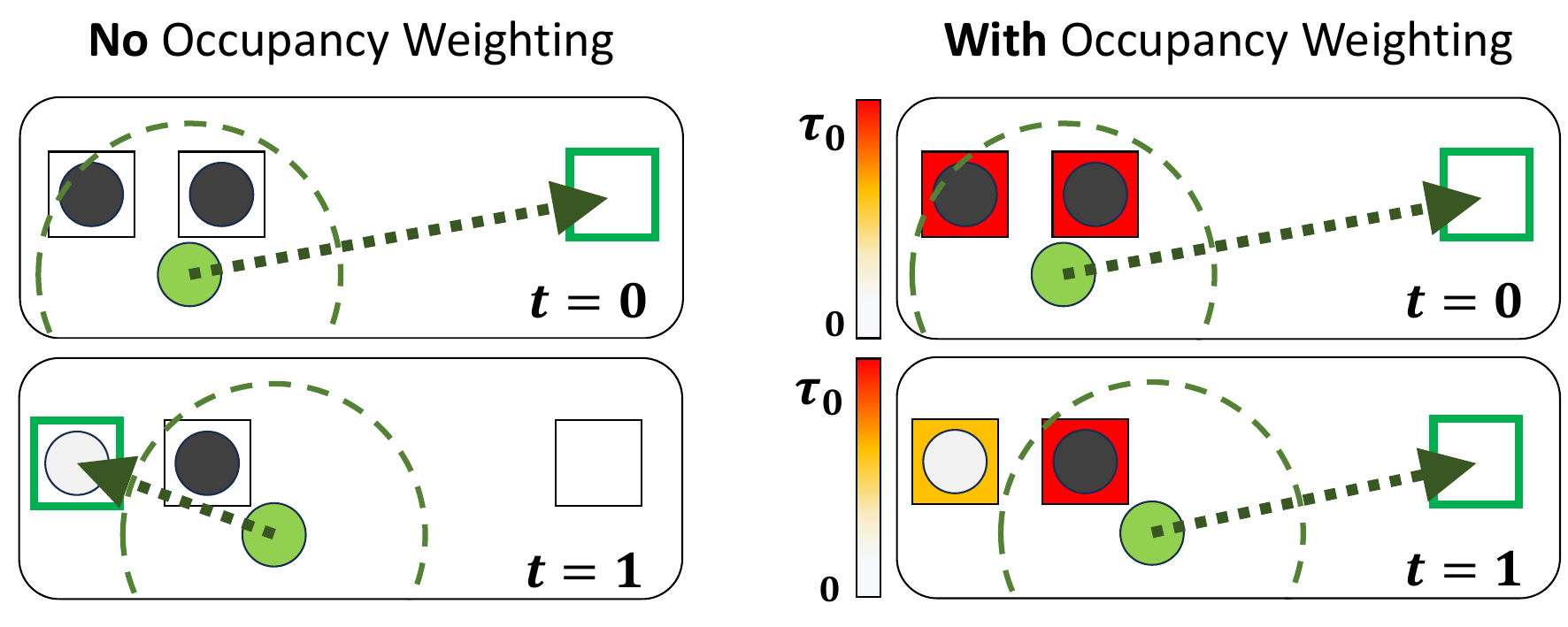}
    \caption{Occupancy Weighting (OW): at $t=0$ the green robot selects the nearest formation point \emph{unoccupied} by its neighbours (gray) to move towards. At $t=1$ the left-most neighbour is out of comms. range and the point it is on appears to be ideal. With OW, the right-most point remains optimal.}
    \label{fig:occupancy-weighting}
\end{figure}

Finally the current position of the robot is transformed into the canonical formation frame, and augmented with a value of $0$ in the third dimension. Robots can then perform a nearest-neighbour search across all three dimensions of the FPs $\Tilde{Q}_i$, which enables them to prefer points with low OW values.
The effect with and without OW is visualised in Figure \ref{fig:occupancy-weighting}. Without OW, a robot becomes stuck oscillating between two FPs as it moves out of communication range with a robot occupying an FP.

\begin{algorithm}[b]
\footnotesize
\caption{For each robot $i$}\label{algo:formation}
\begin{algorithmic}[1]
    \FOR{$j \in \mc{N}(i)$}
        \FOR{$\Tilde{q}_i \in \Tilde{Q}_i$}
            \IF{$\left\| \begin{bmatrix} \mathbf{I}~ \mathbf{0} \end{bmatrix} \Tilde{q}_i -  \leftindex^{\mc{G}}{X}_{i,0} ^{-1}\circ \mathbf{p}_{j,0} \right\|< r_N$}
            \STATE $\Tilde{q}_i(\tau) \leftarrow \tau_0$
            \ENDIF
        \ENDFOR
    \ENDFOR
    \FOR{$\Tilde{q}_i \in \Tilde{Q}_i$}
        \IF{$\left\| \begin{bmatrix} \mathbf{I}~ \mathbf{0} \end{bmatrix} \Tilde{q}_i -  \leftindex^{\mc{G}}{X}_{i,0} ^{-1}\circ \mathbf{p}_{i,0} \right\|< r_C$}
            \STATE $\Tilde{q}_i(\tau) \leftarrow 0$
        \ELSE
            \STATE $\Tilde{q}_i(\tau) \leftarrow \max \left(0,~ \Tilde{q}_i(\tau) - 1 \right)$
    \ENDIF
    \STATE $\mathbf{g}_i = X^{\mc{G}}_{i,0} \circ \begin{bmatrix} \mathbf{I}~ \mathbf{0} \end{bmatrix} \left( \underset{\Tilde{q}_i \in \Tilde{Q}_i}{\arg\min} \left\| \Tilde{q}_i - \begin{bmatrix}\leftindex^{\mc{G}}{X}_{i,0} ^{-1} \circ \mathbf{p}_{i,0} & 0\end{bmatrix}^\top \right\| \right)$
    \ENDFOR
\end{algorithmic} 
\end{algorithm}

\subsection{Experiments: Shape Formation and Continuous Consensus}
Robots of radius $1$~m are initialised with random positions and headings $\boldsymbol{p}_i$ in a $100 \times 100$~m$^2$ environment. They follow a non-holonomic unicycle dynamics model, and plan paths with a time window of $T_H=1.5$~s towards randomly generated goals with a maximum velocity $v_{max}=2$~m/s and a turning radius $r_T=2$~m. Additionally we set $\sigma_u=0.001$, $\sigma_d=0.1$, $\sigma_r=0.01$. In all of our experiments in this work we perform \emph{one} iteration of inter-robot message passing per timestep only, and set $N_I=2$. For all experiments in this work we present results over $50$ trials and a range of letter formations with $d_{min}=r_N=2~$m, $r_S=4~$m, and $\tau_0=10^3$.

In the Global Consensus layer, each robot holds a sliding window of $W=3$ variables, updated every timestep ($T_S=1$). The first of these variables is initialised with a prior factor with observation $\leftindex^{\mc{G}}{x}_{i}^0=\boldsymbol{p}_i$. The strengths of the factors in this layer are set with $\boldsymbol{\sigma_p}=[10, 10, \pi]$, $\boldsymbol{\sigma_s}=0.1\boldsymbol{\sigma_p}$, and $\boldsymbol{\sigma_r}=0.01\boldsymbol{\sigma_p}$.

\subsubsection{Convergence on Formation Parameters}
We vary the radius of communication $r_C$ and the number of robots $N_R$ and measure the number of message passing iterations taken for the swarm to reach a consensus on the formation parameters. We define the convergence criteria to be when the mean inter-robot deviation in the robots' formation parameter beliefs is less than $0.1$m in position, and less than $0.01$~rad in heading.

We compare against the distributed consensus algorithm used in \cite{meanshift}, where agents effectively average their interpretations of formation position and orientation in their local groups. We use the best performing values ($c_1 = c_2 = 1.6$, $\alpha=0.8$) from the work and consider the scenario of a static formation (stationary with respect to time).

\begin{figure}[h]
     \centering
     \begin{subfigure}[t]{0.55\linewidth}
         \centering
         \includegraphics[width=\textwidth]{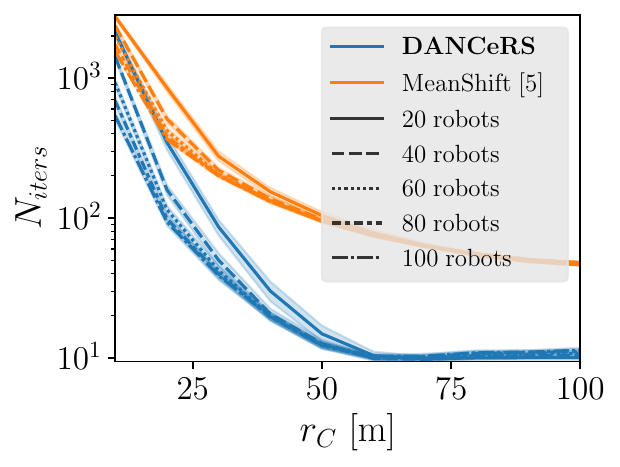}
         \caption{}
         \label{experiment:convergence}
     \end{subfigure}
     \begin{subfigure}[t]{0.42\linewidth}
         \centering
         \includegraphics[width=\textwidth]{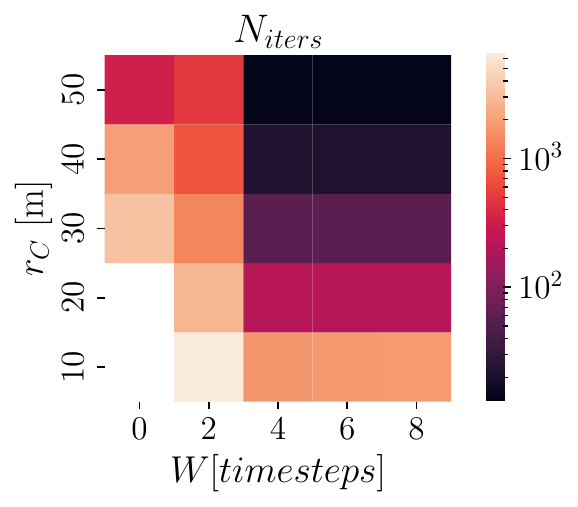}
         \caption{}
         \label{experiment:temporal}
     \end{subfigure}
        \caption{(a): As the communications radius $r_C$ and number of robots $N_R$ increase, it takes fewer iterations for the swarm to converge onto a consensus on the formation parameters.\\
                 (b): As sliding window size $W$ and $r_C$ increase, the number of iterations to convergence decreases, shown for $N_R=30$.}
\end{figure}

Figure \ref{experiment:convergence} shows that we achieve an order of magnitude improvement over the baseline. As $r_C$ increases robots are seen to converge to the global formation parameters in fewer iterations as they are able to exchange information with more robots. This effect is also seen as the number of robots $N_R$ and therefore their density in the environment increases.

\subsubsection{Effect of Sliding Window on Convergence}
We investigate the effect of the length of the sliding window $W$ of variables in the Global Consensus layer on the number of iterations taken until convergence. We consider the case of $N_R=30$ and vary $r_C$.
Figure \ref{experiment:temporal} shows that increasing $W$ results in faster convergence of the formation parameters within the swarm. This trend is more prominent as $r_C$ increases.

When the radius of communication is small, robots forming a local group temporarily negotiate on their beliefs of the formation parameters. However as the variables in GBP are memory-less, when robots leave their local group (altering the factor graph topology) they would lose the previously negotiated belief. Instead using a sliding window of variables, robots are able to maintain their most recent belief due to the new marginalisation factor $f_p$ attached to the now oldest variable.
As shown in Figure \ref{fig:temporal_formation}, the effect of the sliding window is to maintain the belief of the oldest variable but increase its covariance. With a longer sliding window a robot's belief becomes more susceptible to change when it encounters a new clique of robots with a differing belief.

\subsubsection{Shape Formation}
\begin{figure*}[h]
    \centering
     \begin{subfigure}[b]{0.17\linewidth}
         \centering
         \includegraphics[width=\textwidth]{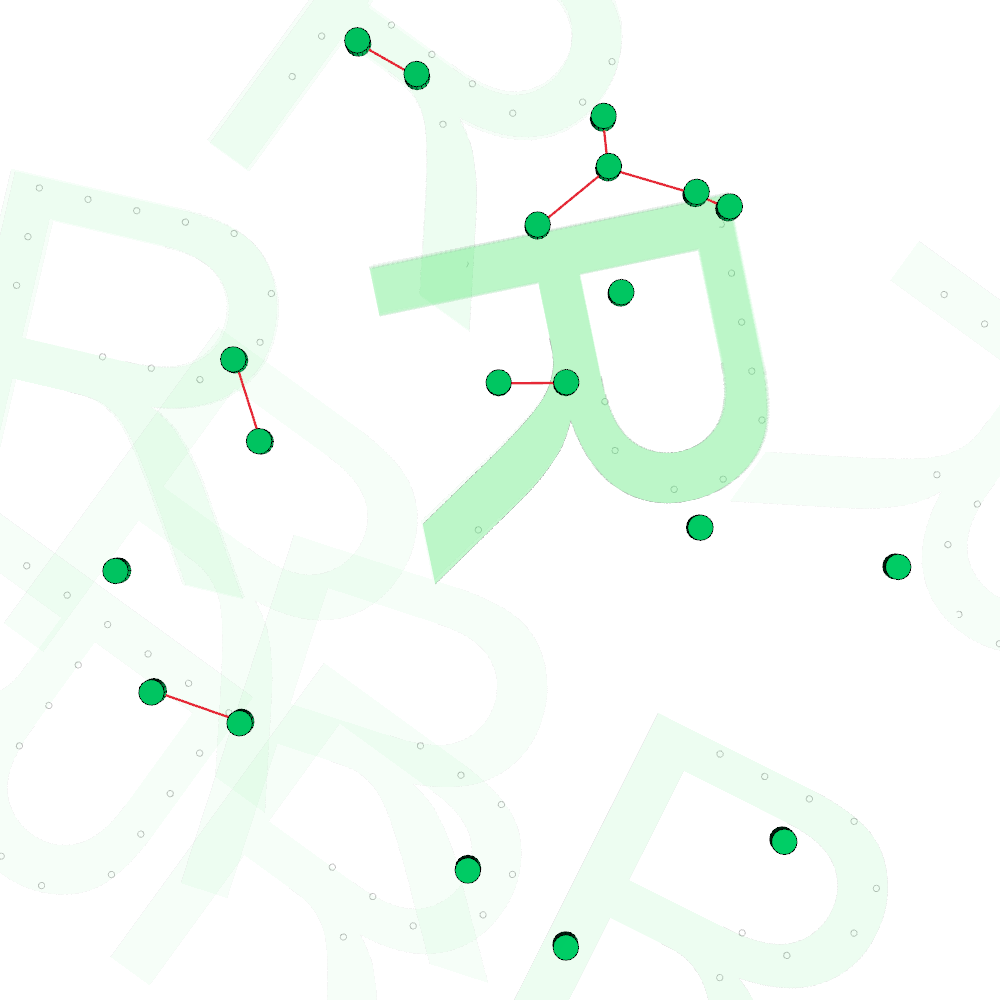}
     \end{subfigure}
     \begin{subfigure}[b]{0.17\linewidth}
         \centering
         \includegraphics[width=\textwidth]{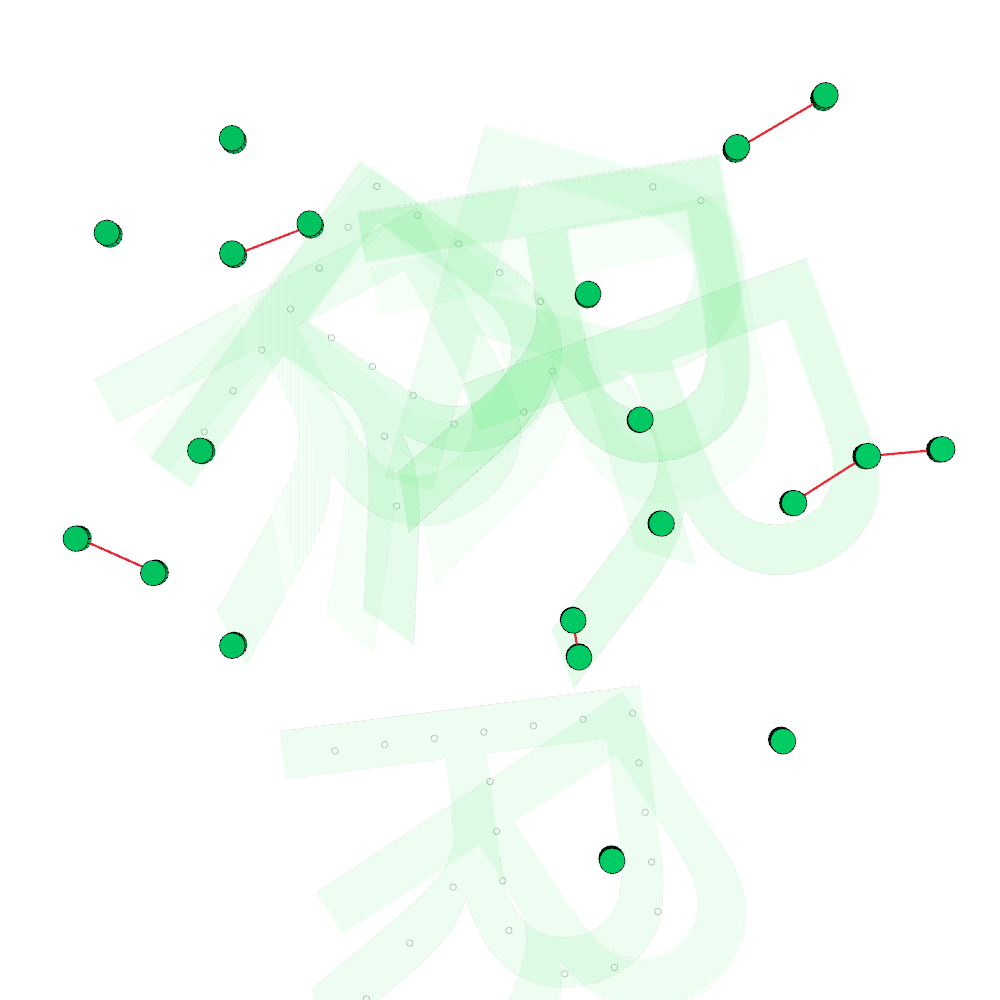}
     \end{subfigure}
     \begin{subfigure}[b]{0.17\linewidth}
         \centering
         \includegraphics[width=\textwidth]{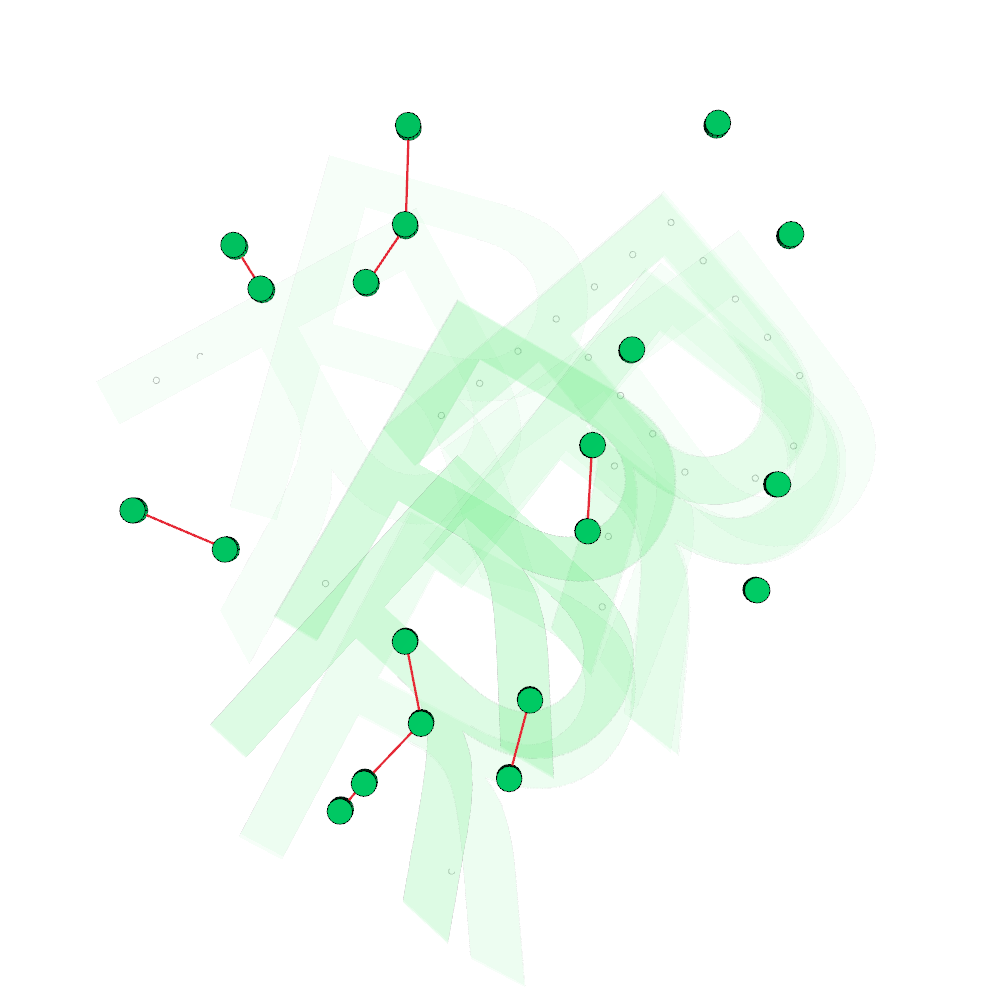}
     \end{subfigure}
     \begin{subfigure}[b]{0.17\linewidth}
         \centering
         \includegraphics[width=\textwidth]{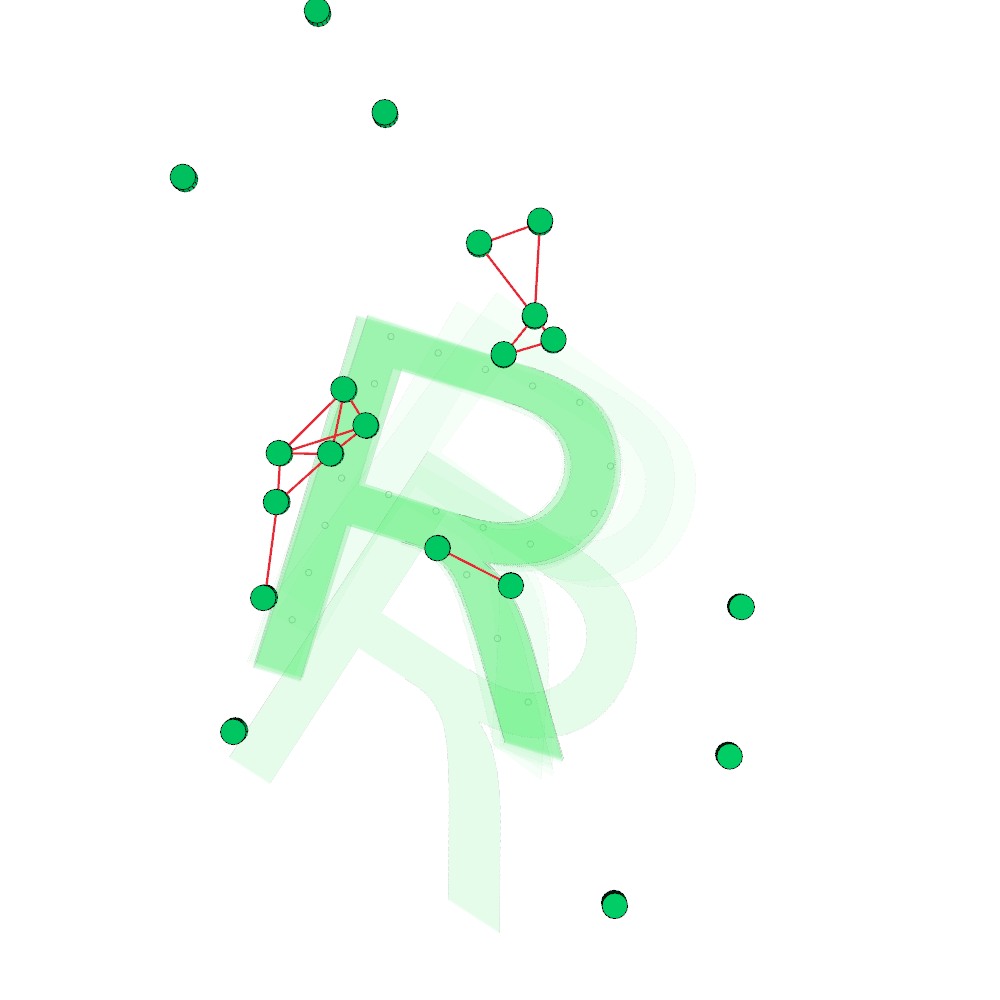}
     \end{subfigure}
     \begin{subfigure}[b]{0.17\linewidth}
         \centering
         \includegraphics[width=\textwidth]{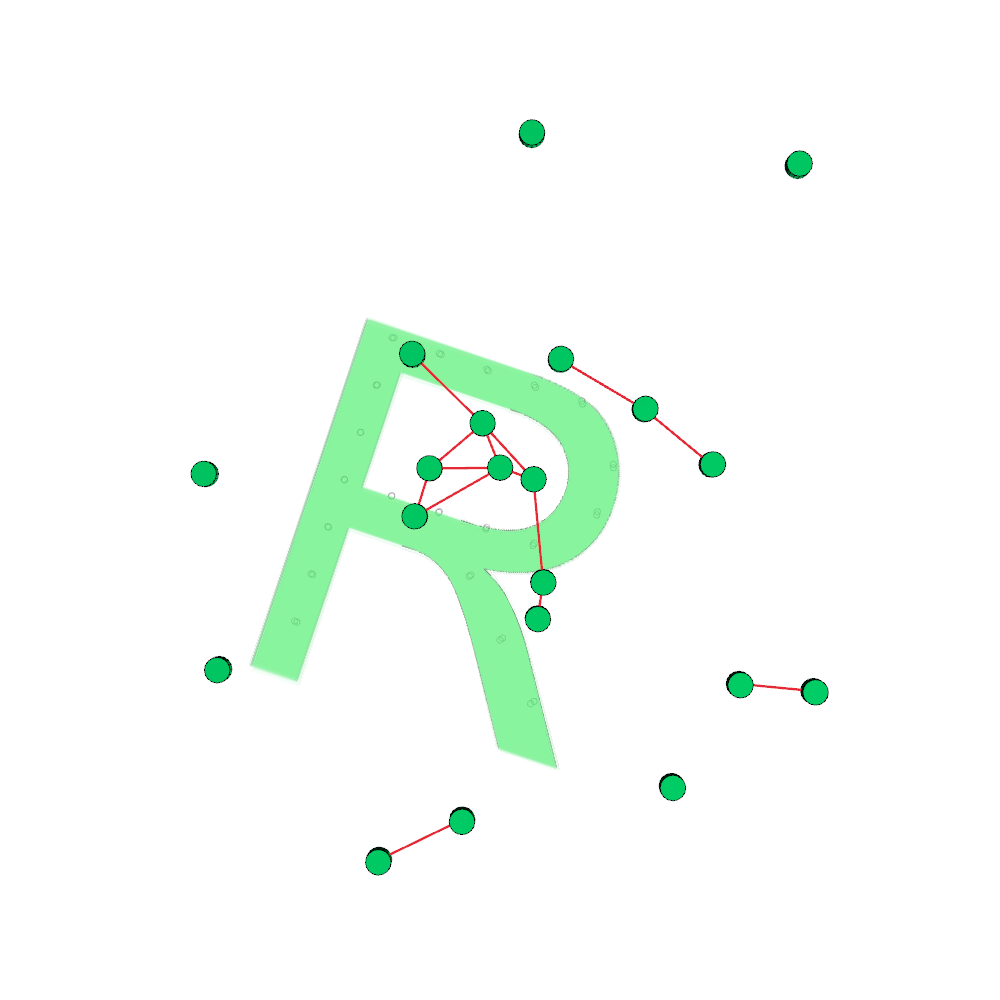}
     \end{subfigure}
        \caption{Robots with a very small comms. radius $r_C$ move randomly across the environment creating a dynamic graph (red lines show connectivity). Due to the sliding window of Global Consensus variables they are able to gradually form a consensus on the formation parameters (visualised with the shaded letter R).}
        \label{fig:temporal_formation}
\end{figure*}

\begin{figure*}[h]
    \centering
     \begin{subfigure}[b]{0.13\linewidth}
         \centering
         \includegraphics[width=\textwidth]{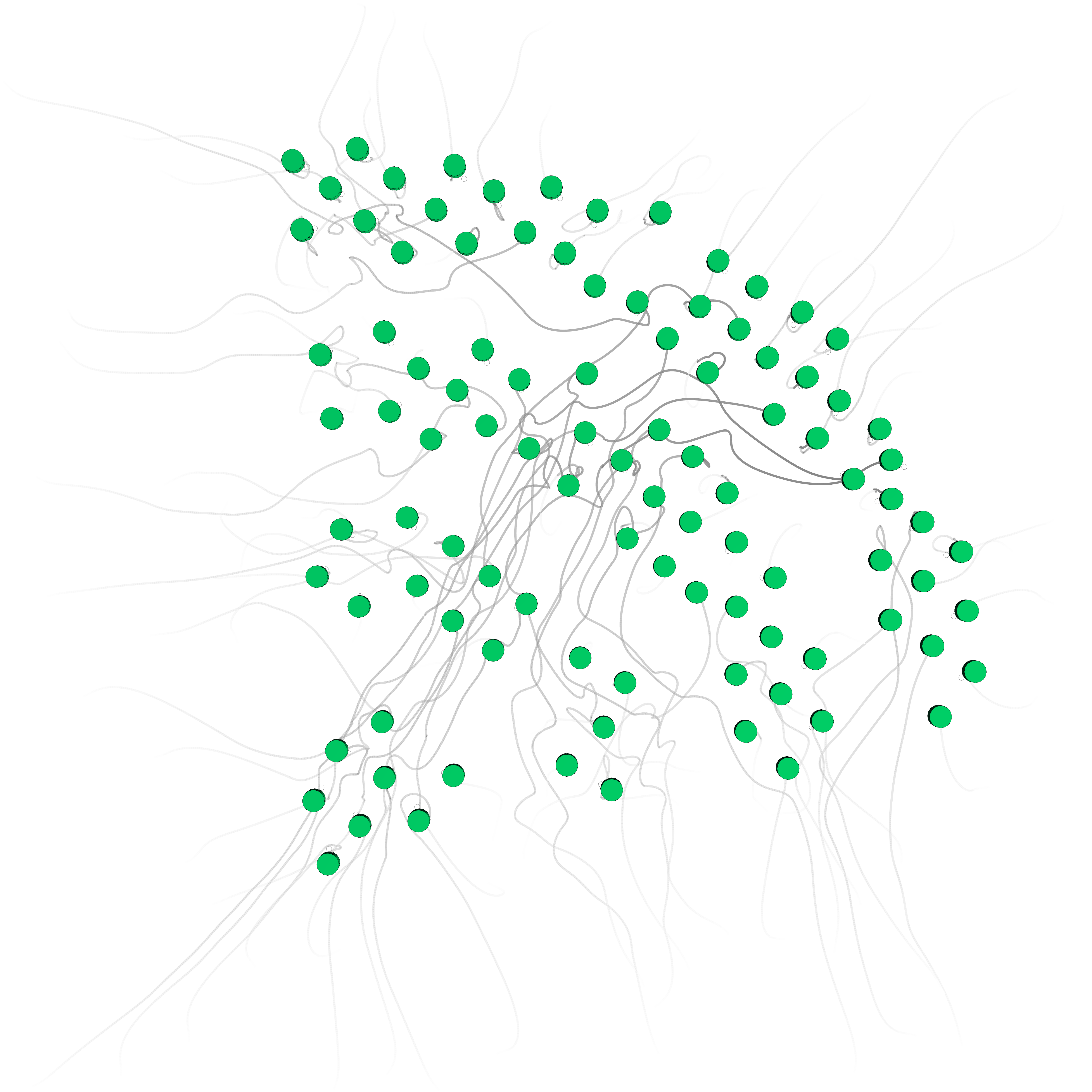}
     \end{subfigure}
     \begin{subfigure}[b]{0.13\linewidth}
         \centering
         \includegraphics[width=\textwidth]{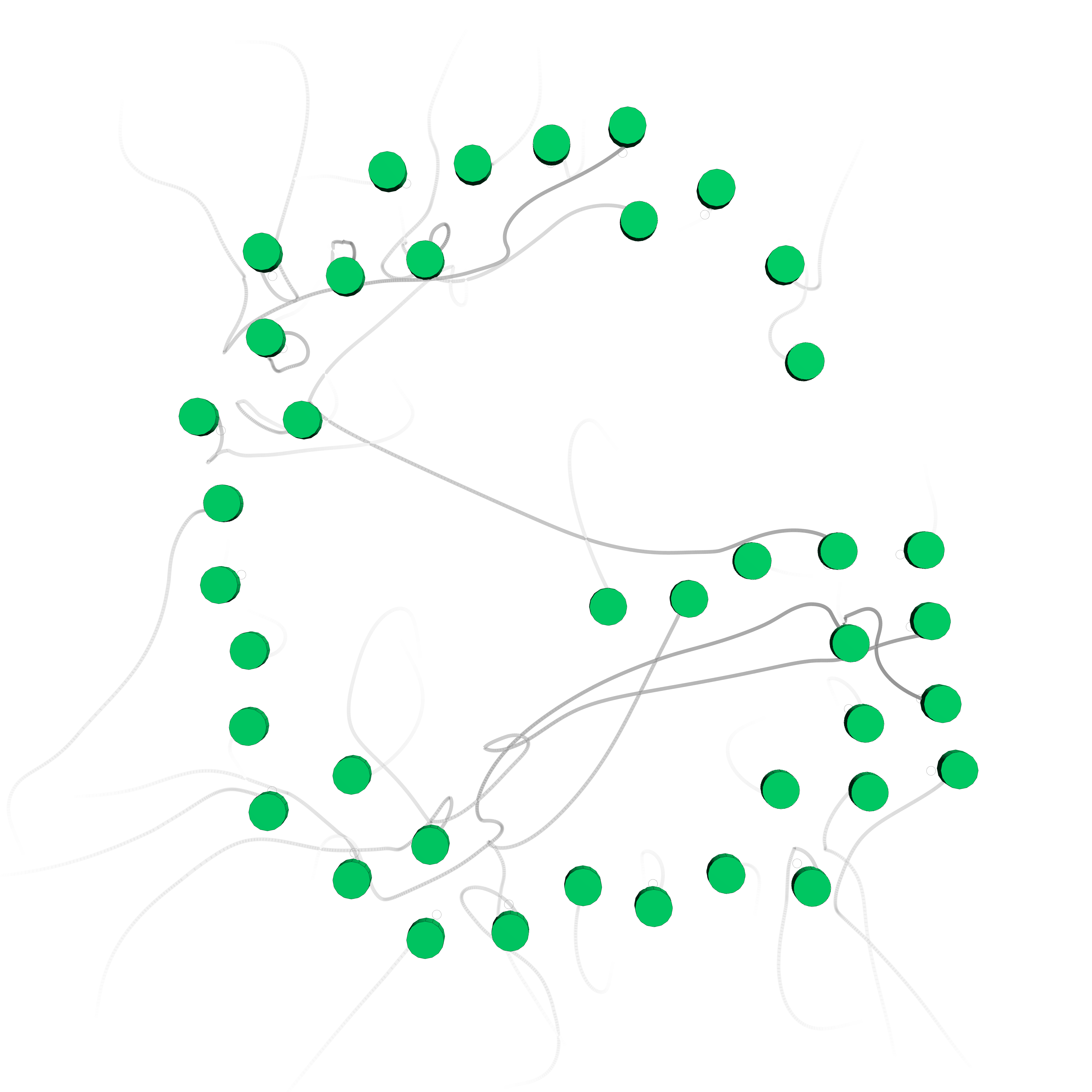}
     \end{subfigure}
     \begin{subfigure}[b]{0.13\linewidth}
         \centering
         \includegraphics[width=\textwidth]{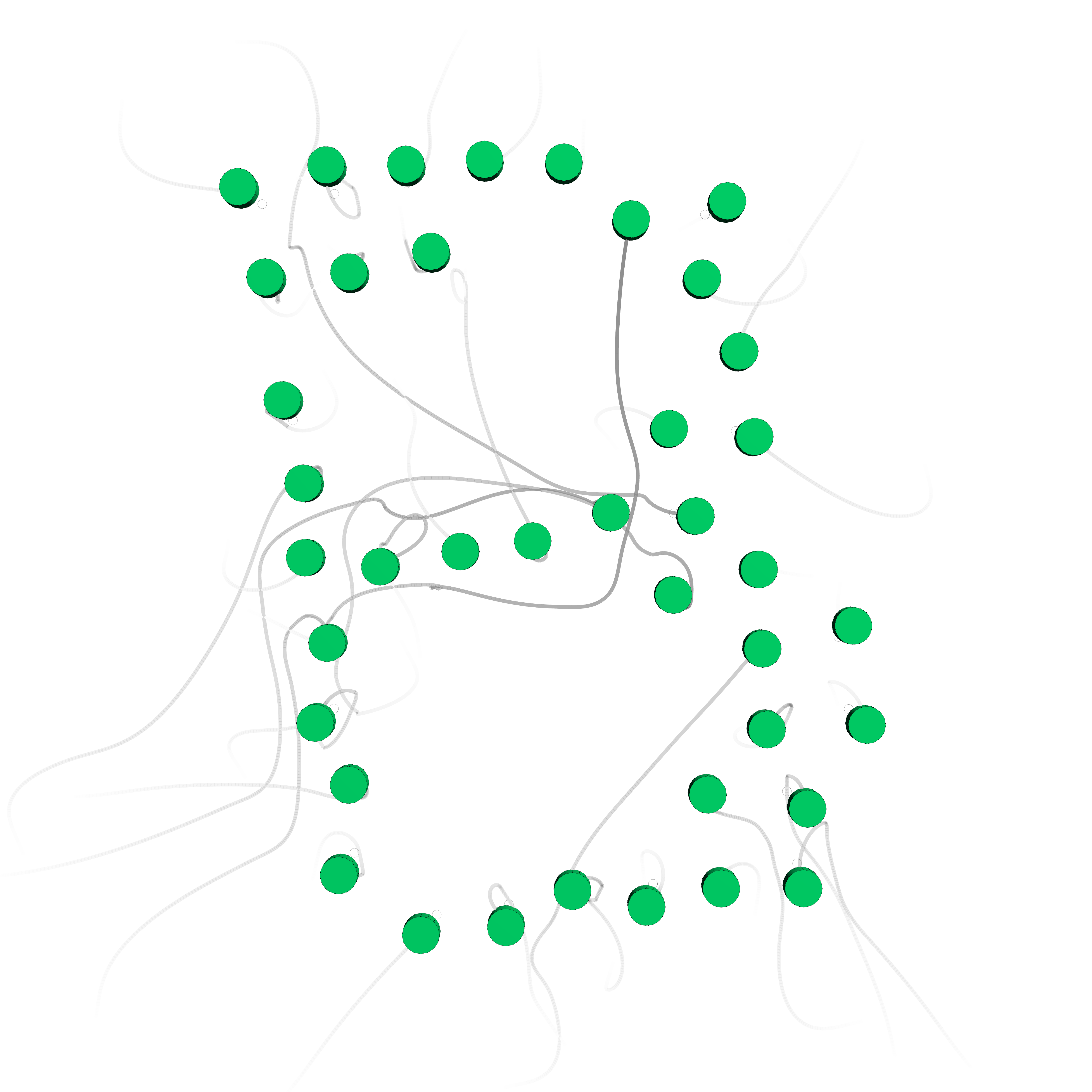}
     \end{subfigure}
     \begin{subfigure}[b]{0.13\linewidth}
         \centering
         \includegraphics[width=\textwidth]{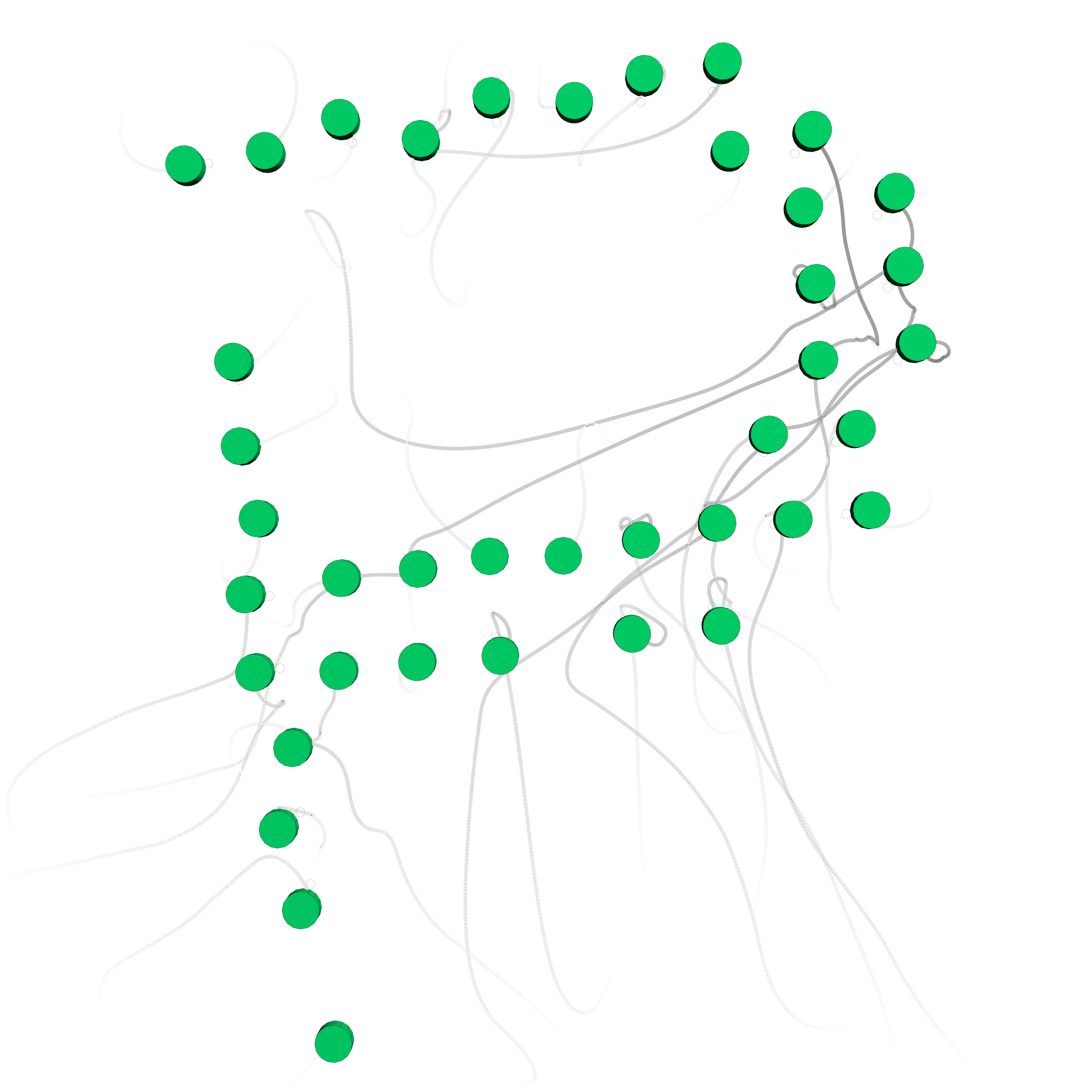}
     \end{subfigure}
     \begin{subfigure}[b]{0.13\linewidth}
         \centering
         \includegraphics[width=\textwidth]{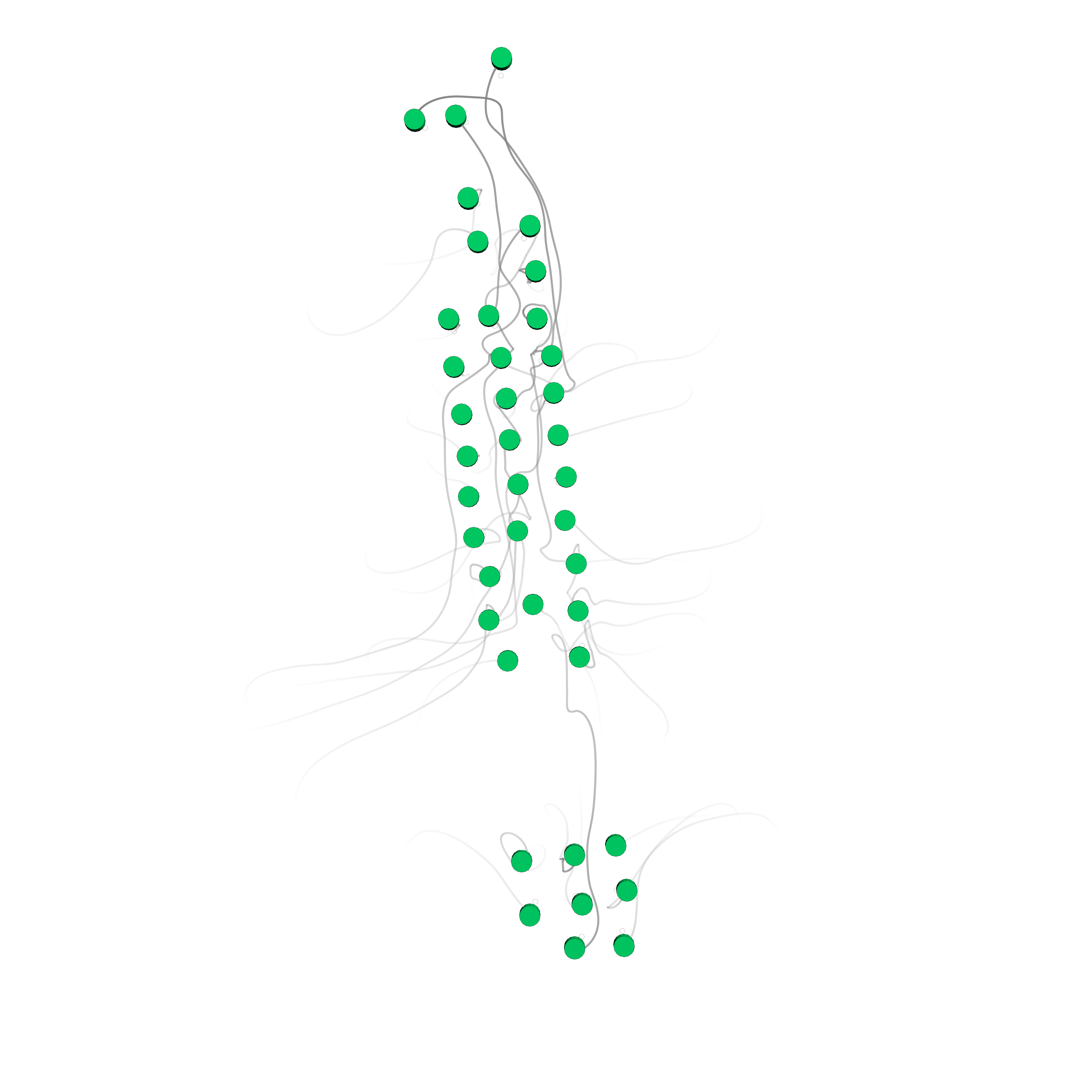}
     \end{subfigure}
        \caption{Qualitative examples of the shapes our method can enable swarms to create. Robot paths are shown in gray. Joint path planning and shape formation is possible even for shapes with multiple disconnected components including `!' and `wifi'.}
        \label{fig:shape_formation}
\end{figure*}

We present qualitative experiments on robots jointly planning paths to form a pre-defined shape whilst achieving consensus on its position and orientation. The stopping criteria for the experiments is when the distance from each point within the formation to its nearest robot is less than a threshold $r_R$~m. We perform experiments over a range of shapes including those in Figures \ref{fig:teaser}, \ref{fig:temporal_formation}, \ref{fig:shape_formation}, and the supplementary video.


Compared to the target-free, mean-shift based shape formation method in \cite{meanshift} which is applicable only to shapes consisting of one connected component, in our method robots explore the shape when they observe that all formation points nearby are occupied by other robots. We are thus able to handle disjoint shapes comprising sparse regions, such as the Smiley Face formation which mean-shift based methods cannot.

\subsection{Consensus over a discrete decision-space}
We now show how the same framework used for the \emph{continuous} consensus problem can be used for reaching consensus over a \emph{discrete} decision space. This problem can be thought of as a variation on the classical \emph{Best-of-N} problem, where a swarm of agents with limited sensing and communication capabilities must reach agreement on a single decision from a discrete set of $N_D$ possible options. This set may represent specific target locations for the swarm to visit, or simply various behaviours for the swarm to exhibit (e.g. `aggregation', `exploration', `formation'). The critical objective is for all agents to agree on the same choice, as consensus within the swarm is essential for coherent collective action. 

\subsubsection{Global Consensus Layer}
We consider that robots only perform optimisation over the Global Consensus layer $\mc{G}$ and do not perform path planning.

Robots are initialised with decisions $X_{i,d} \in [0,N_D-1]$ and prior factors $f_p$ having observation values $z_p = \gamma^{-1}(X_{i,d})$ connected to $\leftindex^{\mc{G}}{X}_{i}$ with a factor strength given by $\sigma_p$.

\subsection{Experiments: Discrete Consensus}
To validate our algorithm we compare against two recent works which propose distributed solutions to the discrete consensus problem, namely an Entropy based consensus algorithm~\cite{consensus_ECA} (which we denote as ECA) and a Probabilistic decision-making consensus algorithm (PCA)~\cite{consensus_PCA}. 
In the former, robots maintain a discrete probability distribution over the $N_D$ options, and exchange with their neighbours their strongest choice, along with the discrete entropy of their probability distribution, representing a certainty about the choice.
In the latter, robots exchange not only their most preferred decision, but also a list of other robots exhibiting the same decision as themselves (their `local consensus group'). The agents then perform internal updating of their states based on the information received from their neighbours.
We choose these algorithms as a baseline for comparison as they were seen to outperform classical consensus methods such as majority rule \cite{majority} and k-unamity \cite{k-unamity}, and demonstrated impressive scalability.
Additionally, we show comparisons against the method in \cite{meanshift} as like our method, it is a consensus algorithm operating on a continuous decision space.

It should be noted that in PCA~\cite{consensus_PCA}, the authors assume a fixed connectivity in the network of robots. They also assume that when a robot leaves or joins a local consensus group (by having changed its exhibited decision), the next iteration of the algorithm only proceeds after this news has reached all other robots. In a real-world network and with dynamically changing graphs this may not be reasonable and may increase the time taken for the swarm to converge to a decision.

Using the initialisation of robots in the experiments from \cite{consensus_PCA}, robots are placed in a random triangular grid structure as in Figure \ref{fig:teaser} (bottom) such that the distance between any pair of robots is at least $5 $m. Robots are initialised with random preferred decisions $X_{d,i}$, and $r_C$ is varied in increments of $6 $m. We define a swarm to have converged to a decision when all agents exhibit the same discrete decision.

\subsubsection{Discrete Convergence}
\begin{figure}[h]
    \centering
    \includegraphics[width=0.8\linewidth]{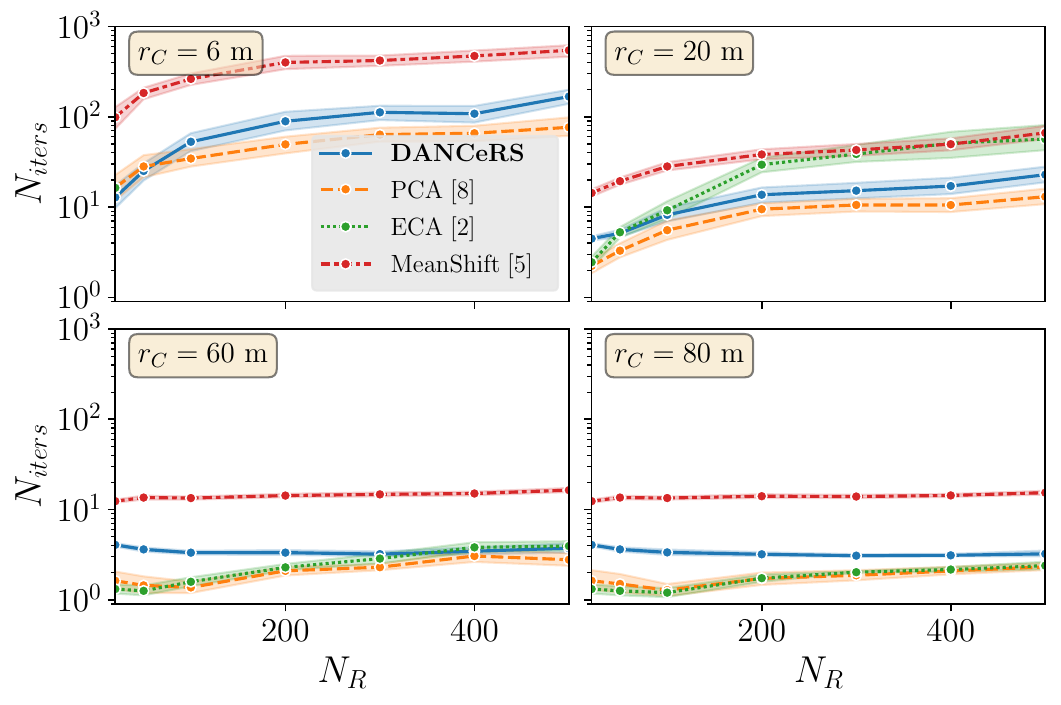}
    \caption{Number of iterations $N_{iters}$ taken for the swarm to converge to the same discrete decision as communication radius $r_C$ and number of robots $N_R$ vary.}
    \label{exp:discrete_consensus}
\end{figure}

We measure the number of message passing iterations $N_{iters}$ until convergence as we vary the communications radius $r_C$ and the number of robots $N_R$. Figure \ref{exp:discrete_consensus} shows that as $r_C$ increases it takes fewer iterations for the swarm to converge onto a common decision. For small $r_C$ it takes longer to converge as the swarm size increases.

We plot only the trials that converged within $1000$ timesteps. Notably when $r_C=6$~m the ECA method failed to converge, and resulted in many local high-certainty decision groups. At higher $r_C$, our method demonstrated its scalability, taking a constant number of iterations to converge as $N_R$ increased.

For the PCA algorithm we followed their assumption that changes to local decision groups are propagated throughout the swarm instantaneously, although in real-world experiments this would vastly increase $N_{iters}$ with $N_R$ as the amount of information shared between robots increases over time as the size of the dominant decision group increases.

\begin{figure}[h]
    \centering
    \includegraphics[width=0.8\linewidth]{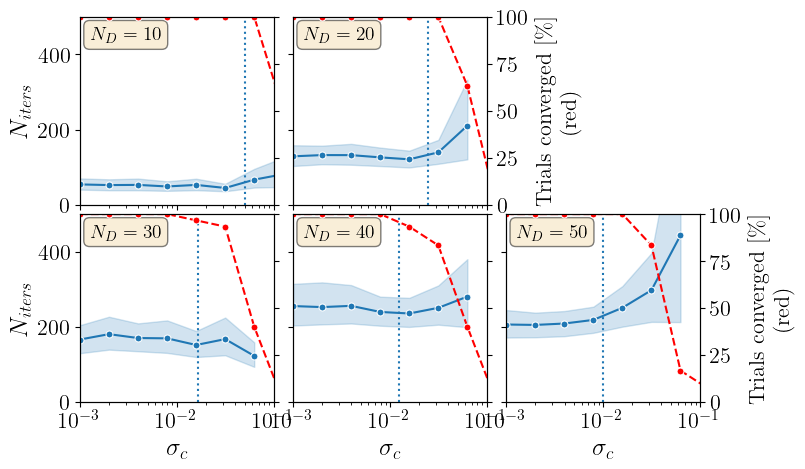}
    \caption{Effect of consensus factor strength $\sigma_c$ and number of decisions $N_D$ on convergence (blue), and \% of total trials that converged (red). Vertical dashed lines show $\sigma_c=\frac{0.5}{N_D}$.}
    \label{exp:discrete_seed}
\end{figure}
\subsubsection{Effect of sigma parameter sweep}
We perform an ablation over the strength $\sigma_c$ of the inter-robot consensus factors $f_c$ and the number of discrete decisions $N_D$. In GBP the `strength' of a factor reflects the extent to which a variable's belief may change during optimisation. We hypothesise that an ideal value of $\sigma_c$ is $\frac{0.5}{N_D}$, as the $\gamma$ partitions the continuous domain $[0,1]$ into $N_D$ discrete bins.

In Figure \ref{exp:discrete_seed} the blue graphs show how the choice of $\sigma_c$ and $N_D$ affect the time to convergence, with the proportion of successful trials (within $1000$ timesteps) shown in red.
Larger values of $\sigma_c$ relate to weaker inter-robot consensus factors, leading to fewer successful trials. The vertical blue dotted lines at $\sigma=\frac{0.5}{N_D}$ indicate a favourable upper bound for the factor strength, where there were very few trials that did not converge.
Smaller values of $\sigma_c$ would result in stronger beliefs after convergence, and would not be suitable for problems where the global consensus parameter is time-varying; this is a promising avenue for future work.

\subsubsection{Informed Robots}
A proportion of the swarm may be `informed' agents, or may have access to salient information about the decision to exhibit. Alternatively it may be desirable to control the large swarm through a small number of agents. We vary the proportion $\zeta$ of such `seed' robots, which all exhibit the same decision and measure the proportion of the swarm that converges to the same decision within $1000$ timesteps. We consider the case of $N_R=500$ and the difficult case of $r_C=6$~m such that each robot in the triangular grid is in communication with its immediate neighbours only.

For our method, we use a strong prior factor on the robot's Global Consensus layer variable with $\sigma_p=10^{-10}$. Table \ref{table:experiment_discrete} shows that our method resulted in convergence of the swarm to the seed robot decision throughout the range of seed robot proportions considered, from $\zeta=0.002$ to $\zeta=0.1$ (corresponding to 1 and 50 seed robots respectively). 

\begin{table}[h]
\centering
\caption{\% Trials converged as proportion of seed robots $\zeta$ varies for $N_R=500$ and $r_C=6$~m.}
\label{table:experiment_discrete}
\begin{tabular}{@{}lrrrrrrr@{}}
\toprule
$\zeta$                   & 0.002  & 0.01  & 0.02  & 0.05  & 0.10  & 0.15  & 0.20   \\ \midrule
ECA~\cite{consensus_ECA}                 &   0 &   0 &   0 &   0 &   0 &   0 &   0  \\
PCA~\cite{consensus_PCA}                 &   9 &  18 &  31 &  94 & 100 & 100 & 100  \\
\textbf{DANCeRS}    &  80 & 100 & 100 & 100 & 100 & 100 & 100  \\ \bottomrule
\end{tabular}%
\end{table}
\section{CONCLUSIONS}
We have presented DANCeRS, a distributed algorithm for achieving consensus in robot swarms using Gaussian Belief Propagation (GBP). By modelling swarm coordination as a factor graph, we enable scalable and decentralised decision-making in both continuous and discrete domains, relying only on peer-to-peer communication. We demonstrated its flexibility through two proxy applications: joint shape formation and path planning, as well as collective decision-making over discrete options, showcasing its ability to handle both types of consensus problems. The lightweight and fully distributed nature of GBP means that our method is well-suited for real-world devices with low power and compute; each robot is only responsible for optimisation over its local portion of the global factor graph, and inter-robot messages from each variable of a robot's GBP stack consist of an $n$-vector and an $n\mathord{\times}n$ symmetric covariance matrix, with $n\mathord{=}3$ for shape formation and $n\mathord{=}1$ for discrete consensus.
Some limitations are that robots must store all points in a static formation, which can be memory-intensive for large shapes, and that discrete consensus assumes a fixed set of options, limiting adaptability to dynamic decision spaces. Despite these, DANCeRS provides a scalable, distributed solution for real-world multi-robot coordination, enabling robots to jointly negotiate continuous and discrete decisions in dynamic environments.



\addtolength{\textheight}{-12cm}   





\bibliographystyle{unsrt} 
\bibliography{bib}

\end{document}